\documentclass[10pt,twocolumn,letterpaper]{article}

\usepackage[pagenumbers]{cvpr} 

\usepackage{booktabs,multirow,siunitx}

\usepackage{colortbl}
\usepackage{arydshln}

\usepackage{listings}
\usepackage{xcolor}

\definecolor{codebg}{rgb}{0.95,0.95,0.95}

\usepackage{enumitem}
\definecolor{cvprblue}{rgb}{0.21,0.49,0.74}
\usepackage[pagebackref,breaklinks,colorlinks,allcolors=cvprblue]{hyperref}

\def\paperID{} 
\def\confName{CVPR}
\def\confYear{2026}

\title{Abstract 3D Perception for Spatial Intelligence in Vision-Language Models}

\author{Yifan Liu$^{1,2}$, Fangneng Zhan$^{2,3}$, Kaichen Zhou$^3$, Yilun Du$^2$, Paul Pu Liang$^3$, Hanspeter Pfister$^{2 \dagger}$ \\
$^1$Tsinghua University \quad
$^2$Harvard University \quad
$^3$Massachusetts Institute of Technology \\
{\tt\small liuyifan22@mails.tsinghua.edu.cn,\quad pfister@seas.harvard.edu}
}

\begin{document}
\maketitle
\begin{abstract}
Vision-language models (VLMs) struggle with 3D-related tasks such as spatial cognition and physical understanding, which are crucial for real-world applications like robotics and embodied agents. We attribute this to a modality gap between the 3D tasks and the 2D training of VLM, which led to inefficient retrieval of 3D information from 2D input. To bridge this gap, we introduce SandboxVLM, a simple yet effective framework that leverages abstract bounding boxes to encode geometric structure for VLMs. Specifically, we design a 3D Sandbox reconstruction and perception pipeline comprising four stages: generating multi-view priors with abstract control, proxy elevation, multi-view voting and clustering, and 3D-aware reasoning. Evaluated in zero-shot settings across multiple benchmarks and VLM backbones, our approach consistently improves spatial intelligence, achieving an 8.3\% gain on SAT Real compared with baseline methods for instance. These results demonstrate that presenting VLMs with 3D abstractions substantially enhances their 3D reasoning ability without additional training, suggesting new possibilities for general-purpose embodied intelligence.

\end{abstract}    
\section{Introduction}
\label{sec:intro}

\begin{figure}[t]
  \centering
  \includegraphics[width=1.0\columnwidth]{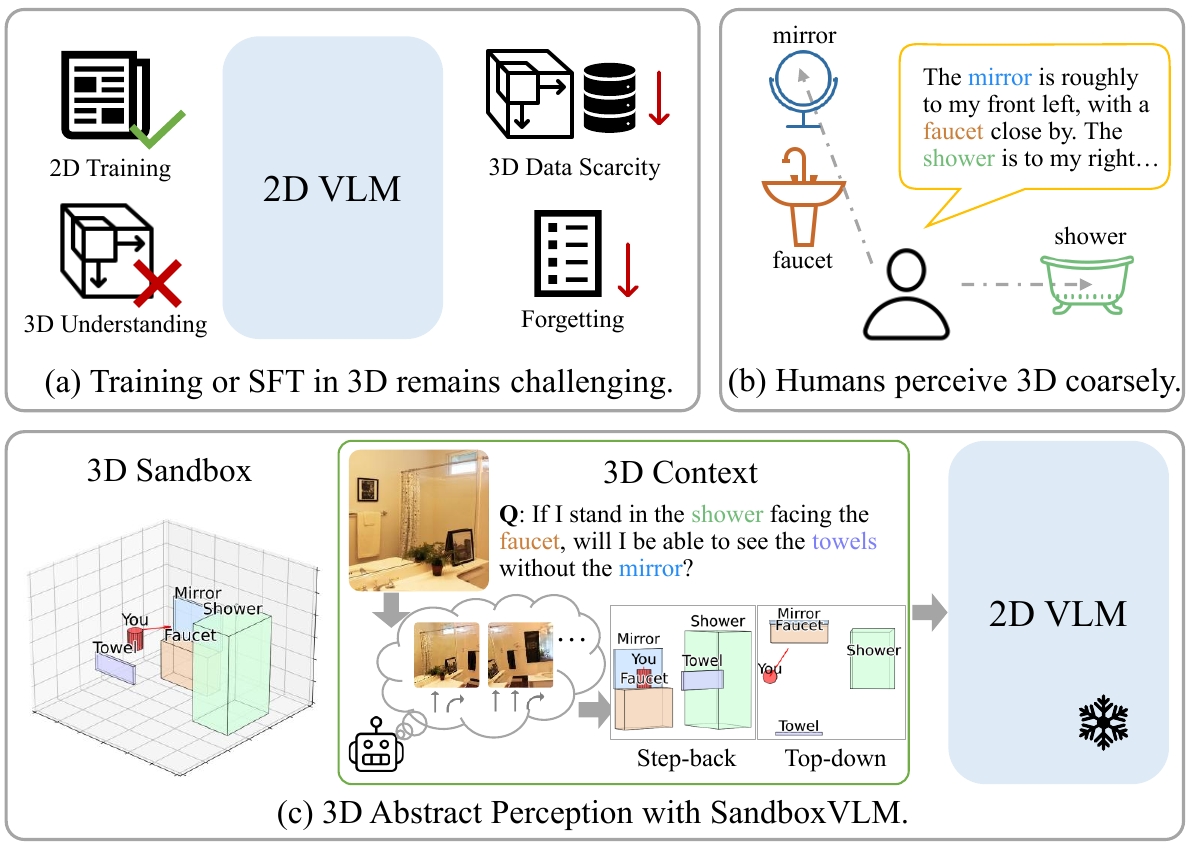}
  \caption{Motivation of SandboxVLM. (a) Existing VLMs are trained without 3D awareness. Training or supervised fine-tuning (SFT) VLMs with 3D suffers from a lack of 3D data and forgetting. (b) Humans, however, reason effectively in 3D through coarse, relational understanding. (c) SandboxVLM follows this principle of abstract perception, providing a coarse but informative 3D context for zero-shot VLM reasoning.}
  \label{fig:header}
  \vspace{-1em}
\end{figure}

Vision-language models (VLMs) such as GPT-5, Gemini, and Qwen3-VL~\cite{bai2025qwen3vl} have recently demonstrated remarkable ability in linking visual and textual information. They can describe scenes, answer questions, and reason compositionally about image content without task-specific post-training. Yet despite their impressive breadth, these models remain fundamentally two-dimensional: they interpret the world as projections rather than as volumetric, physical spaces. Trained almost entirely on 2D images and 1D text~\cite{liu2024llava}, they lack grounding and understanding in the inherently 3D nature of the real world~\cite{yang2025thinking}. As a result, they struggle with tasks that require genuine spatial understanding, such as reasoning under viewpoint changes, estimating relative positions, or predicting outcomes of object interactions~\cite{chow2025physbench}. These limitations prevent current VLMs from achieving robust spatial understanding, a cornerstone of embodied intelligence and robotics.


Recent efforts have begun to extend vision-language models (VLMs) toward understanding 3D scenes. Early approaches, such as 3D-LLM~\cite{hong20233d}, Cube-LLM~\cite{cho2024cube}, and ShapeLLM~\cite{qi2024shapellm}, inject point-cloud or multi-view features into 2D VLMs to improve geometric reasoning. While effective in constrained settings, these methods depend heavily on dense 3D supervision, curated datasets, or domain-specific architectures, limiting their scalability as general-purpose models (Fig.~\ref{fig:header}(a)).
Moreover, these training-based methods can only be applied to open-sourced models, and cannot leverage the strong and ever-evolving proprietary VLMs such as GPT-5.
More recently, models like MindJourney~\cite{yang2025mindjourney} and world models~\cite{zhou2025svc,yang2024hunyuan3d} enrich 2D perception with implicit 3D or temporal priors with video diffusion or generative modeling, but they still operate over 2D or sequential representations. Consequently, even state-of-the-art VLMs struggle to achieve human-like spatial reasoning.

Humans, however, perceive and act effectively in 3D space without constructing metrically precise models of the world~\cite{wu2018human0, hills2008human1,dubois2021human2}. Our spatial understanding is inherently \textit{abstract}: we grasp relative positions, directions, and interactions through coarse perception rather than detailed reconstruction, as shown in Fig.~\ref{fig:header}(b). We do not estimate millimeter-level distances, yet we can effortlessly catch a ball or navigate a crowded room. This observation suggests that intelligent 3D reasoning does not necessarily require full geometric recovery---only an abstract structural understanding of the scene. Inspired by this principle of human \textit{abstract perception}, we ask: can a VLM reason about 3D space by relying on minimal, symbolic abstractions rather than dense 3D supervision?

We propose \textit{SandboxVLM}, a simple yet effective framework that injects 3D structural information into existing VLMs (Fig.~\ref{fig:header}(c)). Instead of reconstructing dense geometry, SandboxVLM represents each scene with a compact set of abstract 3D bounding boxes that encode spatial arrangement and physical dynamics while discarding low-level visual details. This design is directly inspired by the principle of \textit{abstract perception}: retaining only the coarse structural cues necessary for reasoning while leaving appearance information to the original image. Through a lightweight proxy elevation process, 2D cues are lifted into 3D and organized into an interpretable, symbolic scene representation that existing VLMs can reason over.

Specifically, given a 2D image, we first equip scene understanding with common-sense priors extracted from a video diffusion model~\cite{zhou2025svc}, generating a sequence of multi-view observations(Sec.~\ref{sec:mv_priors}). We then estimate per-frame depth using an off-the-shelf model~\cite{wang2025vggt}. The VLM identifies task-relevant objects, which serve as prompts for a 2D segmentation model~\cite{ravi2024sam2}. The masked objects are then lifted into 3D via our Proxy Elevation Module (Sec.~\ref{sec:proxy}) and structured into abstract bounding boxes through a Multi-view Voting Clustering (MVC) algorithm (Sec.~\ref{sec:mv_voting}). Finally, the abstract 3D scene is rendered from informative viewpoints and fed back into the VLM for spatial and physical reasoning (Sec.~\ref{sec:3d_reasoning}).

Despite its simplicity, this abstraction exposes crucial spatial relationships, enabling VLMs to reason about relative position and motion without post-training or explicit 3D data. Evaluated in zero-shot settings across multiple benchmarks, SandboxVLM consistently enhances 3D reasoning performance and outperforms state-of-the-art methods by a large margin (e.g., $+3.8\%$ on SAT-Real~\cite{ray2024sat}). These results demonstrate that a coarse, symbolic notion of structure can substantially improve 3D spatial understanding. Our contributions are threefold:
\vspace{0.3em}
\begin{itemize}
\item We introduce the concept of \textbf{abstract perception}: a human-inspired perspective that approaches 3D reasoning through coarse structural cues rather than precise geometric reconstruction.
\item We present \textbf{SandboxVLM}, a training-free framework that injects symbolic 3D structure into existing VLMs via proxy elevation and bounding-box representations, enabling 3D reasoning without requiring dense supervision or architectural changes.
\item We demonstrate that such abstraction substantially improves zero-shot spatial reasoning across benchmarks and backbone models, providing a generalizable plug-and-play solution to enhance spatial intelligence for the ever-evolving VLM community.
\end{itemize}
\vspace{0.3em}

We hope SandboxVLM motivates further exploration of human-inspired geometric priors and abstract structural representations as a foundation for scalable embodied intelligence in VLMs.

\section{Related Work}
\label{sec:related}

\noindent\textbf{Vision-Language Models.}
Recent large vision--language models (VLMs), such as Qwen3-VL~\cite{bai2025qwen3vl} and GPT-5, demonstrate impressive multimodal understanding of images, text, and video. Yet the majority of these models remain fundamentally grounded in 1D/2D representations (text and images), without an understanding of 3D representations. To address this gap, a new wave of 3D-LLMs has emerged~\cite{hong20233d,tang2024minigpt,cho2024cube,qi2024shapellm,xiong20253ur}.
3D-LLM~\cite{hong20233d} constructs point-level features from multi-view images and feeds them into 2D VLM backbones with additional 3D localization tokens.
MiniGPT-3D~\cite{tang2024minigpt} aligns point clouds with LLMs by leveraging pretrained 2D VLM priors and a cascaded four-stage training strategy with mixture-of-query-experts for efficient 3D-language alignment.
ShapeLLM~\cite{qi2024shapellm} equips large language models with 3D object understanding by integrating a point-cloud encoder with an LLM and performing 3D visual instruction tuning to support multimodal dialogue and embodied interaction tasks.
These works collectively highlight a rapid trend toward integrating geometric and physical reasoning within vision-language models. Nevertheless, most still depend on heavy 3D supervision, large-scale 3D datasets, or domain-specific architectures that limit their compatibility with general-purpose VLMs. 

\begin{figure*}[t]
  \centering
  \includegraphics[width=0.95\textwidth]{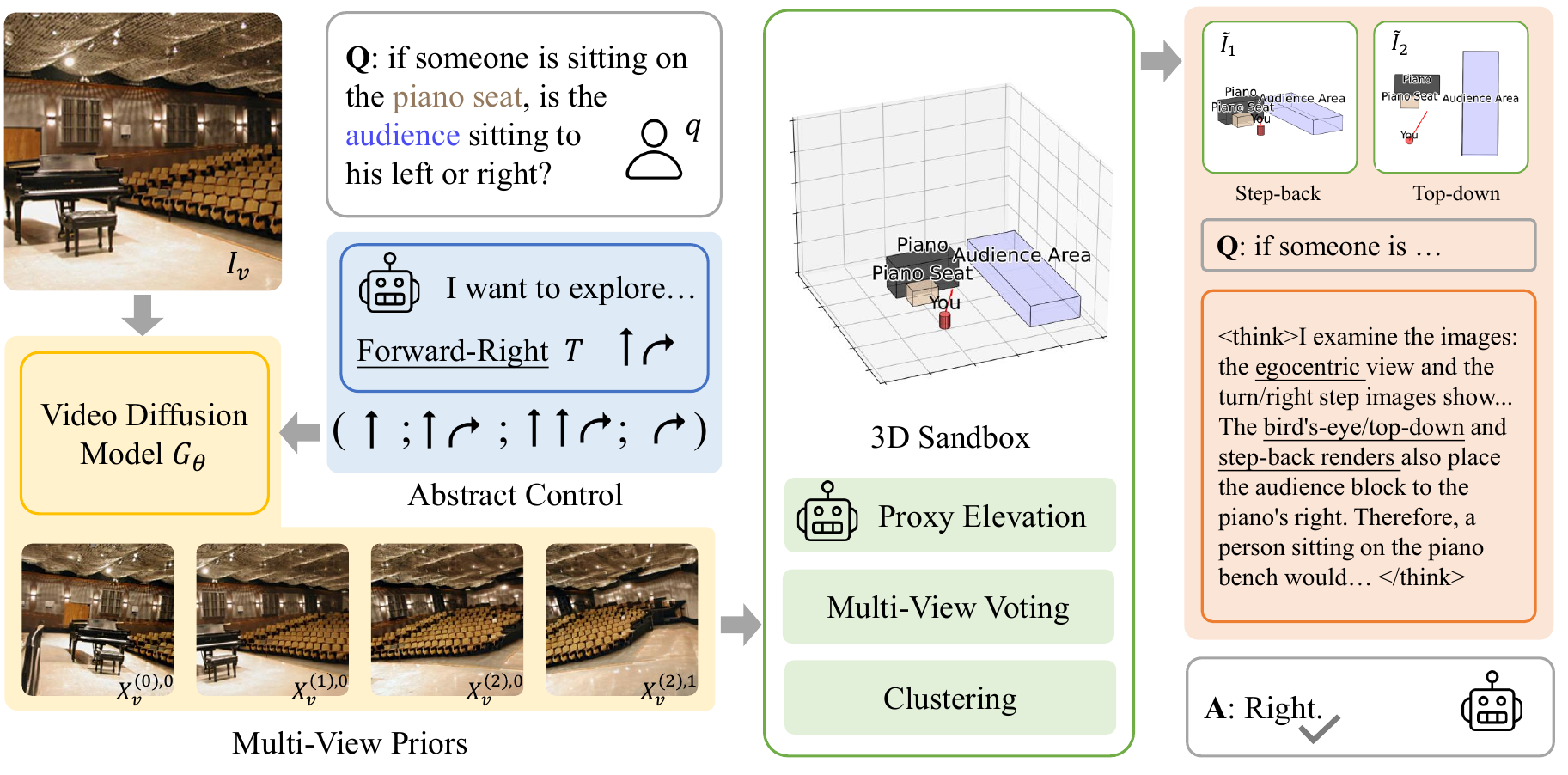}
  \caption{Overview of the SandboxVLM pipeline. Given an input image and a textual query, the system builds a compact, 3D-aware, query-conditioned context for a vision-language model (VLM). A video diffusion prior first expands the input into a short multi-view sequence along imagined trajectories guided by abstract control provided by the VLM. Inside the 3D Sandbox module, an off-the-shelf depth estimator predicts per-frame depth and camera parameters, while the VLM identifies task-relevant objects that guide a 2D segmenter to produce instance masks. The masked regions are lifted into coarse 3D proxies and merged across views through a Multi-View Voting and Clustering step to form abstract 3D bounding boxes. Finally, informative renderings of these 3D abstractions are composed with the query and fed back into the VLM for spatial and physical reasoning.}
  \label{fig:pipeline}
  \vspace{-1em}
\end{figure*}

\noindent\textbf{3D Representations and Spatial Reasoning.}
A broad line of research has explored 3D scene representations, ranging from mesh-based~\cite{wu2020pqnet}, voxel-based~\cite{choy20163d,shi2020pvrcnn}, and point-cloud models~\cite{qi2017pointnet,wang2025vggt,zhou2025page4ddisentangledposegeometry} to neural implicit representations such as NeRF~\cite{mildenhall2021nerf} and its many variants~\cite{barron2021mipnerf,park2021nerfies,zhou2023dynpoint}.
These methods capture detailed geometry and appearance but typically rely on dense multi-view supervision and precise camera poses, which are often impractical for real-world agents.
Another line of work models 3D scenes at a higher level of abstraction, representing them as graphs of objects and relations~\cite{armeni20193d,wu2025universalscenegraph} or as cognitive maps~\cite{yang2024thinking,yin2025spatial} that encode spatial layouts for reasoning.
APC~\cite{lee2025perspectiveawarereasoningvisionlanguagemodels} employs abstract perspective change to enable perspective-aware reasoning for VLMs.
Recently, VAGEN~\cite{wang2025vagen} introduced Chain-of-Thought prompting into VLMs to enhance spatial intelligence, while world models~\cite{zhou2025svc,yang2024hunyuan3d} and MindJourney~\cite{yang2025mindjourney} integrate 3D or video priors into generative and multimodal frameworks for richer scene understanding.
However, these approaches remain limited to 1D or 2D representations and thus cannot fully represent the 3D world.
In contrast, our proposed \textit{3D Sandbox} representation introduces abstract 3D bounding boxes via a lightweight 2D-to-3D proxy unprojection, enabling VLMs to reason over disentangled spatial cues while avoiding low-level visual distractions and heavy 3D point cloud perception.

\section{Method}
\label{sec:method}

\subsection{Overview}

\textbf{Task Definition.}
We aim to enhance zero-shot spatial reasoning in large vision-language models (VLMs). Given one or more RGB images of a 3D scene $\mathcal{I}{=}\{I_v\}_{v=0}^{V-1}$ (each $I_v{\in}\mathbb{R}^{H\times W\times 3}$, v indexes different views) and a natural-language query $q$, the objective is to answer $q$ by understanding 3D relationships among scene objects. The output is an answer $a$ to the query.

\noindent\textbf{Framework.}
As shown in Fig.~\ref{fig:pipeline}, given an input image $I_v$ (or image set $\mathcal{I}$) and a query $q$, our goal is to construct a compact, 3D-aware, query-conditioned context for a VLM $M_{\psi}$. A video diffusion prior $G_{\theta}$ first expands $I_v$ into short multi-view sequences $\{X_v^{(m),t}\}$ along abstractly-controlled imagined trajectories (Sec.~\ref{sec:mv_priors}). An off-the-shelf model $D_{\theta}$ predicts per-frame depth and camera parameters, while $M_{\psi}$ proposes task-relevant object categories $\hat{\mathcal{O}}$ that prompt a 2D segmenter $S_{\theta}$ to produce masks $\{\mathbf{M}_{v,i}\}$ (Sec.~\ref{sec:proxy}). Masked pixels are lifted into coarse 3D proxies $\mathcal{P}_i$, which a Multi-View Voting Clustering step merges into abstract 3D boxes $\mathcal{B}{=}\{\mathbf{b}_i\}$ (Sec.~\ref{sec:mv_voting}). Informative renderings $\{\tilde{I}_k\}$ of $\mathcal{B}$ are finally composed with $q$ and fed back into $M_{\psi}$ for spatial and physical reasoning, yielding the answer $a$ (Sec.~\ref{sec:3d_reasoning}).

\subsection{Multi-View Priors with Abstract Control}
\label{sec:mv_priors}
The input 2D image $I_v$ provides limited information about the 3D scene, causing 3D ambiguity and making it challenging for VLMs to perform accurate spatial reasoning.
To provide the system with comprehensive 3D information from a single image $I_v$, we utilize a video diffusion prior $G_{\theta}$~\cite{zhou2025svc} to generate a short multi-view video $\{X_v^{(m),t}\}_{t=0}^{T-1}$ that simulates camera motion an imagined trajectory~\cite{yang2025mindjourney}, where $(m)$ denotes the index of the trajectory, and $t$ stands for the timestep inside a trajectory. To focus the computation of video diffusion on task-related objects and directions, similar to how humans mentally explore promising viewpoints during spatial reasoning~\cite{wu2018human0, hills2008human1,dubois2021human2}, we design an abstract control mechanism that guides the trajectory generation based on the input query $q$. Specifically, we process $q$ and $I_v$ with a VLM $M_{\psi}$, inquiring about the most relevant direction with the task among a predefined set of abstract camera motions: 
\begin{equation}
    \mathcal{T}=\{\text{left, fwd-left, fwd, fwd-right, right}\}.
\end{equation}
The selected abstract motion $c^*\in\mathcal{T}$ is then instantiated as $M$ candidate camera trajectories, each a sequence of transforms whose viewpoints coarsely follow the desired direction $\{\hat{\mathbf{T}}_{v}^{(m),t}\}_{t=0}^{T-1}$ for $m{=}0,\dots,M-1$, where $v$ indexes the input view, $m$ indexes the candidate trajectory, and $t$ indexes the time step (frame) within that trajectory. For example, if $c^*{=}\text{fwd-left}$, the trajectories sweep from turning left to moving forward, covering intermediate directions. We condition the video diffusion model $G_{\theta}$ on each trajectory to generate the corresponding multi-view sequence:
\begin{equation}
    \{X_{v}^{(m),t}\}_{t=0}^{T-1} \;=\; G_{\theta}\!\left(I_v, \{\hat{\mathbf{T}}_{v}^{(m),t}\}_{t=0}^{T-1}\right).
\end{equation}
In the following equations, $m,t$ will be omitted for clarity. Optionally, we collect all synthesized views for $I_v$ as $\mathcal{X}_v \,{=}\, \bigcup_{m=1}^{M}\{X_{v}^{(m),t}\}_{t=0}^{T-1}$, providing comprehensive multi-view observations that are rich in 3D information.
This abstract control allows the model to focus on generating views that are most informative for answering the query, enhancing the relevance of the multi-view observations for subsequent 3D reasoning.

\begin{figure}[t]
  \centering
  \includegraphics[width=1.0\columnwidth]{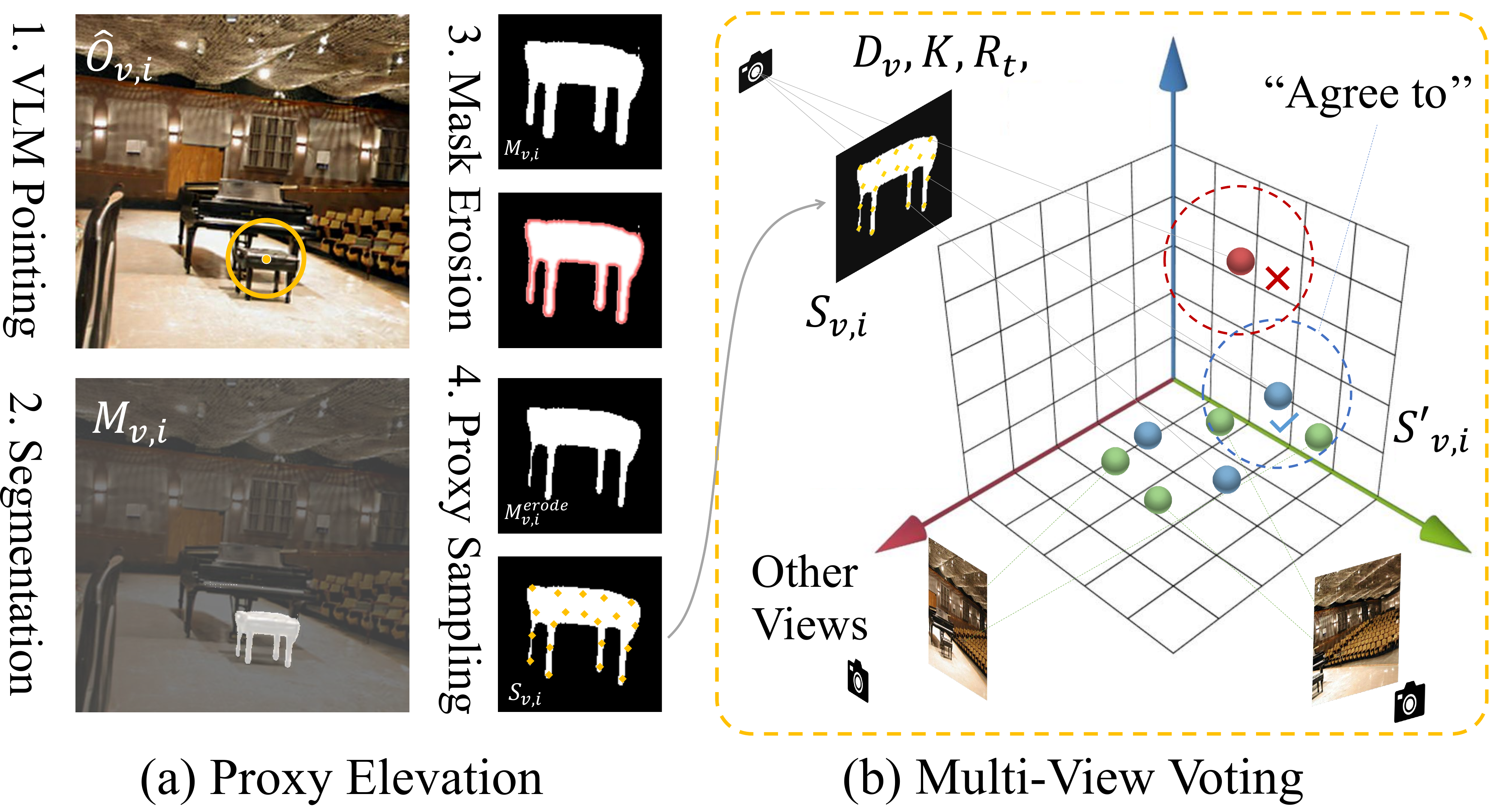}
  \caption{Core modules of 3D Sandbox. (a) Proxy Elevation: The VLM identifies task-relevant objects and their approximate locations. A segmentation model produces object masks, followed by mask erosion and farthest point sampling to select interior proxy pixels. (b) Multi-View Voting: The proxies are unprojected into 3D space and aggregated across views through a cross-view consistency check (“Agree to”) to filter unreliable points. The remaining proxies will be clustered into boxes.}
  \label{fig:method}
  \vspace{-1em}
\end{figure}

\subsection{Proxy Elevation}
\label{sec:proxy}
To aggregate 3D information from the synthesized multi-view images, we propose a Proxy Elevation module that identifies task-relevant objects in each view, segments them, and lifts them into 3D space as proxy points. The process is illustrated in Fig.~\ref{fig:method}(a). These proxies serve as the foundation for constructing a compact 3D abstract representation of the scene. Instead of reconstructing dense visual appearance like point cloud, NeRF~\cite{mildenhall2021nerf} or 3D Gaussian Splatting~\cite{kerbl20233dgaussian}, we focus on extracting sparse yet informative 3D proxies that capture the spatial locations of key objects involved in the query $q$.
To achieve this, we first leverage the VLM $M_{\psi}$ to analyze the input query $q$ with $I_v$ and identify a set of relevant object categories $\hat{\mathcal{O}}{=}\{\hat{o}_i\}$ that are likely to be involved in answering $q$, leveraging the common sense and 2D VQA abilities innate to VLMs. For each category, it also provides a central point coordinate for the instance in the image:
\begin{equation}
    \{\hat{O}_{v,i}\} = M_{\psi}(I_v, q),
\end{equation}
\begin{equation}
    \hat{O}_{v,i} = (\hat{o}_i, [x_i, y_i]),
\end{equation}
where $(x_i, y_i)$ denotes the pixel coordinates of the center of object $o_i$ in $I_v$. Then, they are inputted into a 2D segmentation model $S_{\theta}$~\cite{ravi2024sam2} to generate comparatively precise binary masks $\mathbf{M}_{v,i}{\in}\{0,1\}^{H\times W}$ for each object in each synthesized view:
\begin{equation}
    \mathbf{M}_{v,i} = S_{\theta}(X_v, \hat{O}_i).
\end{equation}

For each mask $\mathbf{M}_{v,i}$, we further prepare it for 3D lifting by sampling pixels within the mask region to represent the object, termed as a 2D proxy. Since mask and depth estimation can be noisy especially on the object's edge, we apply a morphological mask erosion operation on $\mathbf{M}_{v,i}$ to obtain a refined mask $\mathbf{M}_{v,i}^{\text{erode}}$ that focuses on the interior points of the mask. Then we apply Farthest Point Sampling (FPS) on the eroded mask to select a fixed number of pixels (i.e., 30 per object per view) as the 2D proxy points $\mathcal{S}_{v,i}$:
\begin{equation}
    \mathcal{S}_{v,i} = \text{FPS}(\mathbf{M}_{v,i}^{\text{erode}}, N_{\text{pts}}).
\end{equation}
The selected 2D proxy points $\mathcal{S}_{v,i}$ represent the spatial extent of object $o_i$ in view $X_v$.

Finally, we estimate depth maps $\{D_v\}$, camera intrinsics $\mathbf{K}$ and extrinsics $R_t$ for each synthesized view using an off-the-shelf depth estimation model $D_{\theta}$~\cite{wang2025vggt}:
\begin{equation}
    D_v, K, R_t = D_{\theta}(X_v),
\end{equation}
Using the estimated depth maps, the estimated camera intrinsics $\mathbf{K}$ and extrinsics $R_t$ from $D_\theta$, we lift each 2D proxy point $\mathbf{u}{=}(x,y){\in}\mathcal{S}_{v,i}$ into 3D space via back-projection, obtaining $\mathcal{S'}_{v,i}$. The 3D proxy points sketch the spatial extent of the object from the current view.

\begin{table*}[t]
\centering
\caption{Quantitative comparison with three categories of baselines across four benchmarks evaluating spatial reasoning. Our SandboxVLM achieves the highest overall performance, particularly on SAT-Real and PhysBench, demonstrating strong generalization in real-world spatial and physical tasks.}
\label{tab:main}
\setlength{\tabcolsep}{3pt}
\begin{tabular}{l|c:cccc|c}
\toprule
Method & Spatial-Avg & BLINK-Spatial & BLINK-Depth & EmbSpatial & SAT-Real & PhysBench \\
\midrule
\underline{\textit{Generalist VLMs}}&&&&&& \\
GPT-4o &71.0&78.3&79.0&68.3&58.3&50.3\\
GPT-5-mini &78.5&81.8&\underline{82.3}&74.3&75.4&47.1\\
Claude-Sonnet-4 &74.0&80.4&75.8&64.3&75.3&- \\
Gemini-2.5-Pro &80.3&84.6&79.0&\underline{78.4}&79.3&- \\
Qwen2.5-VL-32B~\cite{Qwen2.5-VL} &77.5&\underline{85.3}&77.4&72.3&74.9&43.1\\
Qwen2.5-VL-72B~\cite{Qwen2.5-VL} &71.1&78.3&74.2&73.3&58.7&46.6\\
\midrule
\underline{\textit{Training-Based Models}}&&&&&& \\
VeBrain-8B~\cite{luo2025vebrain} &72.0&81.1&78.2&70.5&58.0&- \\
Magma-8B~\cite{yang2025magma} &66.9&66.4&65.3&64.6&71.3&- \\
Robix-32B-Base~\cite{fang2025robixunifiedmodelrobot} &-&-&-&\textbf{79.0}&79.6&-\\
RoboBrain2.0-32B~\cite{baairobobrainteam2025robobrain20technicalreport} &\underline{81.0}&\textbf{87.4}&79.8&76.6&\underline{80.3}&-\\
Cosmos-Reason1~\cite{nvidia2025cosmosreason1physicalcommonsense} &65.6&73.4&63.7&64.6&60.7&-\\
\midrule
\underline{\textit{Test-time Scaling VLMs}}&&&&&& \\
MindJourney~\cite{yang2025mindjourney} &79.1&81.8&\textbf{83.1}&74.7&78.7&\underline{54.9}\\
\rowcolor{gray!15}
\textbf{SandboxVLM(Ours)}      &\textbf{81.4}&83.7&\underline{82.3}&75.4&\textbf{84.1}&\textbf{58.3}\\
\bottomrule
\end{tabular}
\end{table*}

\subsection{Multi-View Voting and Clustering}
\label{sec:mv_voting}
To obtain high-quality 3D abstract boxes for each object, we propose a Multi-View Voting and Clustering algorithm, depicted in Fig.~\ref{fig:method}(b), that aggregates the lifted 3D proxy points from multiple views. The key idea is to leverage the consensus among different views to robustly decide the real points belonging to the object, and fit an oriented 3D bounding box accordingly. Specifically, for each object $o_i$, we collect all the lifted 3D proxy points from all views $\{\mathcal{S'}^{(m),t}_{v,i}\}$, where $m,t$ denotes the trajectory index and frame in trajectory respectively. 

We define ``Agree to" as a relationship between a 3D point $\mathbf{p}\in\mathcal{S'}^{(m_0),t_0}_{v_0,i_0}$ and a view $X_{v}^{(m),t}$ ($v\neq v_0$):
\begin{equation}
    \text{Agree}(\mathbf{p}, X_{v}^{(m),t}) = 
    \begin{cases}
        1, & \text{if } \exists\,\mathbf{p}'\in\mathcal{S'}^{(m),t}_{v,i_0}\\
           & \quad\text{s.t.}\;\|\mathbf{p}' - \mathbf{p}\|_2 < \delta; \\
        0, & \text{otherwise},
    \end{cases}
\end{equation}
where $\delta$ is a distance threshold. A point $\mathbf{p}$ is considered reliable if it is agreed upon by N views. This operation helps filter out noisy points from depth errors or mask deficiencies that are not consistently observed across multiple views.

After lifting proxy pixels to 3D and filtering for cross-view consistency and outliers, we cluster the remaining points per category using DBSCAN to separate object instances (e.g., multiple chairs). For each cluster, we fit an oriented 3D bounding box via a PCA-based OBB: principal axes are obtained from the covariance eigenvectors, and box extents come from the min/max of points in this PCA frame; the center is the midpoint mapped back to world space. The resulting set of instance boxes $B = {b_i}$ forms a compact, 3D-aware “Sandbox” representation that captures only task-relevant spatial structure for downstream reasoning.

\subsection{3D-Aware Reasoning}
\label{sec:3d_reasoning}
We aim to provide the VLM $M_{\psi}$ with a compact yet informative 3D-aware context that enables accurate spatial reasoning to answer the query $q$. To this end, we render the abstract 3D boxes $\mathcal{B}$ from a set of informative viewpoints that maximize the visibility and coverage of the objects involved in $q$. Specifically, we select these two viewpoints: (1) Step-back view: a view that steps back 2 meters from the original camera to capture the overall spatial arrangement of objects; (2) Top-down view: a bird's-eye view that reveals the horizontal layout of the scene. We compose the rendered images $\{\tilde{I}_k\}$ from these viewpoints along with the query $q$ and original input images $\{I_v\}$ into a final question-answering prompt for the VLM. Given this 3D-aware context, the VLM is allowed to do textual reasoning before answering the query:
\begin{quote}
  \footnotesize\texttt{<thinking> The reasoning. </thinking>  
  <answer> The final answer. </answer>}
\end{quote}
The details of the prompt design and answer formatting are provided in the supplementary material.
\section{Experiments}
\label{sec:experiments}
\begin{table*}[t]
\centering
\caption{Quantitative results on the SAT-Real and SAT-Synth benchmarks. Each block compares baseline, MindJourney, and our SandboxVLM under the same backbone (GPT-4o, GPT-5-mini, or GPT-5). Metrics are reported across five spatial reasoning dimensions: EgoMovement (EgoM), ObjectMovement (ObjectM), GoalAiming (GoalAim), ActionConsequence (ActCons), and PerspectiveTaking (Perspect). SandboxVLM consistently achieves the best or second-best results.}
\label{tab:sat-results}
\setlength{\tabcolsep}{1.4pt}
\begin{tabular}{l *{12}{c}}
\toprule
\multirow{2}{*}{Method} & \multicolumn{6}{c}{SAT Real} & \multicolumn{6}{c}{SAT Synth} \\
\cmidrule(lr){2-7} \cmidrule(lr){8-13}
 & Average & EgoM & ObjectM & GoalAim &ActCons  & Perspect & Average & EgoM & ObjectM & GoalAim & ActCons & Perspect \\
\midrule
\underline{\textit{GPT-4o}} &&&&&&&&&&&& \\
Baseline & 60.3 & 56.5 & 69.3 & 64.0 & 50.0 & 54.5 & 59.2 & 68.8 & 87.5 & 85.7 & 45.5 & 31.6 \\
MindJourney & 69.4 & 78.3 & 60.9 & 70.6 & 78.4 & 57.6 & 61.2 & 87.5 & \underline{81.3} & \underline{92.9} & 45.5 & 26.3 \\
\rowcolor{gray!15}
\textbf{SandboxVLM}      & 77.7 & 91.3 & 66.7 & 84.6 & 82.9 & 62.5 & 66.3 & \underline{81.3} & 68.8 & 85.7 & 60.6 & 47.4 \\
\midrule
\underline{\textit{GPT-5-mini}} &&&&&&&&&&&& \\
Baseline &75.4&82.6&73.9&76.5&81.1&63.6&65.3&\textbf{100.0}&62.5&78.6&48.5&57.9\\
MindJourney &78.7&87.0&73.9&85.3&81.1&66.7&71.4&\textbf{100.0}&62.5&85.7&\textbf{60.6}&63.1\\
\rowcolor{gray!15}
\textbf{SandboxVLM}      & \underline{84.1} & \textbf{100.0} & \textbf{82.6} & \textbf{92.9} & 79.4 & 70.0 & \textbf{75.4} & 93.8 & \underline{81.3} & 85.7 & 63.6 & \underline{68.4} \\
\midrule
\underline{\textit{GPT-5}} &&&&&&&&&&&& \\
Baseline &80.1&82.6&\textbf{82.6}&73.5&\textbf{89.2}&\underline{72.7}&72.5&\textbf{100.0}&\underline{81.3}&\textbf{100.0}&48.5&63.2\\
MindJourney &72.8&82.6&67.4&75.2&78.3&60.6&69.4&87.5&75.0&85.7&\underline{57.5}&57.9\\
\rowcolor{gray!15}
\textbf{SandboxVLM}      & \textbf{84.3} & \textbf{100.0} & \textbf{82.6} & \underline{85.7} & \underline{83.9} & \textbf{73.3} & \underline{73.5} & 93.8 & \textbf{87.5} & \underline{92.9} & 48.5 & \textbf{73.7} \\
\bottomrule
\end{tabular}
\end{table*}
\subsection{Experimental Setting}
\noindent\textbf{Datasets.} We tested our method with visual question answering (VQA) datasets. We evaluate on three datasets for spatial intelligence: SAT (Spatial Aptitude Training Dataset)~\cite{ray2024sat}, BLINK~\cite{fu2024blink} (Spatial-Relation and Relative-Depth splits), and EmbSpatial Bench~\cite{du2024embspatialbench}. For the SAT dataset, we report results on the two splits: Real (test) and Synthetic (validation). The synthetic dataset is subsampled to keep the API calls manageable, and the subsampled dataset will be reported in the Supplementary.
These benchmarks test various aspects of 3D spatial reasoning, including relative position, relative depth, object movement, ego movement, perspective shift, and consequence prediction.
For physics understanding, we assess on the PhysicBench~\cite{chow2025physbench}, which further contains questions about physical interactions and intuitive kinematics. The details of these datasets are provided in the Supplementary.

\noindent\textbf{Baselines.} We compare SandboxVLM against several categories of baselines:
(1)~\textit{Proprietary or open-source generalist VLMs}: GPT-4o, GPT-5-mini, Claude-Sonnet-4, Gemini-2.5-Pro, Qwen2.5-VL-32B and Qwen2.5-VL-72B~\cite{Qwen2.5-VL}.
(2)~\textit{Training-based methods}: VeBrain-8B~\cite{luo2025vebrain}, Magma-8B~\cite{yang2025magma},
Robix-32B-Base~\cite{fang2025robixunifiedmodelrobot}, RoboBrain2.0-32B~\cite{baairobobrainteam2025robobrain20technicalreport}, and
Cosmos-Reason1~\cite{nvidia2025cosmosreason1physicalcommonsense}, 
(3)~\textit{Other test-time scaling method}: MindJourney~\cite{yang2025mindjourney}. 

\noindent\textbf{Implementation.} The hyperparameters ($M, T, N, \delta$, etc.) and framework details are introduced in the Supplementary.

\subsection{Results}
We report the main results in Tab.~\ref{tab:main} for all datasets across baselines to verify the effectiveness of our method, and in Tab.~\ref{tab:sat-results} for the SAT benchmarks with varied backbone VLMs for our model for detailed analysis.

\begin{table*}[t]
\centering
\caption{Ablation study across multiple spatial reasoning benchmarks. We evaluate eight model variants described in Sec.~\ref{sec:ablation}, each isolating a key design choice. The full model achieves the highest overall accuracy, demonstrating that 3D Sandbox is one effective way of modeling spatial structure for VLMs.}
\label{tab:ablation}
\setlength{\tabcolsep}{6pt}
\begin{tabular}{c l *{6}{c}}
\toprule
No. & Method & Average & EgoM & ObjectM & GoalAim & ActCons & Perspect \\
\midrule
(1) & Vanilla VLM & 75.4& 82.6& 73.9& 76.5& \textbf{81.1}& 63.6\\
(2) & Scene-Graph Text Prompt & 77.0& 75.6& 78.3& 76.5& 78.3& \textbf{75.7}\\ 
(3) & Multi-View Images Only & 78.7& 87.0& 73.9& \underline{85.3}& \textbf{81.1}& 66.7\\
(4) & Rendered Point Clouds & 73.7& 82.6& 63.0& 82.4& 78.3& 60.6\\
(5) & 3D Coordinate Text Prompt & \underline{80.8}& \underline{95.6}& \textbf{85.7}& \underline{85.3}& 77.1& 66.6 \\
(6) & Rendered Proxy Points & 77.0& 90.9& \underline{82.6}& 82.4& 71.8& 63.6\\
(7) & Single Image Sandbox & 77.6& \textbf{100.0}& 69.6& 78.6& 77.1& 66.7\\
\rowcolor{gray!15}
(8) & Full SandboxVLM    & \textbf{84.1} & \textbf{100.0}  & \underline{82.6}  & \textbf{92.9}  & \underline{79.4}  & \underline{70.0}   \\
\bottomrule
\end{tabular}
\end{table*}

\noindent\textbf{Main results.}
As shown in Tab.~\ref{tab:main}, SandboxVLM achieves superior average performance across a diverse set of spatial intelligence and physical reasoning benchmarks, consistently outperforming all baseline categories. Specifically, it attains an average accuracy of 81.4\% on spatial benchmarks, surpassing the strong generalist baseline GPT-5-mini by 2.9\%. Remarkably, SandboxVLM also exceeds the performance of models explicitly fine-tuned for spatial understanding, such as RoboBrain2.0-32B, highlighting the effectiveness of our test-time scaling paradigm. These results validate the advantage of scaling approaches that leverage the latent capabilities of evolving large VLMs without requiring additional costly retraining.
On PhysBench, SandboxVLM achieves the highest performance of 58.3\%, outperforming the strongest baseline of MindJourney~\cite{yang2025mindjourney} by 3.4\%. This further demonstrates the strength of our 3D spatial reasoning framework in capturing intuitive aspects of physical understanding.

While the performance on BLINK and EmbSpatialBench is relatively lower than that of training-based baselines, this gap can be attributed to the simpler question styles of these datasets, where extensive task-specific training offers a distinct advantage. Nevertheless, the consistently strong results across all spatial benchmarks highlight the robustness and generalization ability of our approach in enhancing large VLMs' 3D spatial reasoning capabilities.

\noindent\textbf{Results on the SAT Benchmark.}
As shown in Tab.~\ref{tab:sat-results}, we further evaluate SandboxVLM across the subtasks of the SAT dataset. Using GPT-4o as the backbone and tested on SAT-Real, SandboxVLM achieves a remarkable 17.4\% improvement in average accuracy over the GPT-4o baseline. It also surpasses the test-time scaling model MindJourney~\cite{yang2025mindjourney} by 8.3\% on average on SAT-Real, demonstrating the advantage of abstracting spatial information into a unified 3D representation rather than relying on a sequence of generated 2D images.

We further analyze the performance of SandboxVLM with different backbone VLMs on the SAT benchmarks. 
As the backbone scales from GPT-4o to GPT-5-mini and GPT-5, performance consistently improves across all subsets, demonstrating that our framework generalizes well to evolving foundation models and effectively enhances their 3D spatial reasoning.
Specifically, applying our method to GPT-4o not only significantly boosts its overall performance but also narrows the gap with one of the most advanced VLMs today, GPT-5 (77.7\% vs 80.1\%) in the SAT Real benchmark. 
When applied to GPT-5, our method continues to yield a 4.2\% improvement over the strong backbone. 
These results indicate that our framework provides stable and scalable gains across backbone generations.
Importantly, even as VLMs evolve and scale rapidly, our approach remains effective, suggesting its potential for long-term applicability and influence on future generations of large vision-language models, without the need for costly retraining or fine-tuning. 

\noindent\textbf{Error analysis.} Error propagation is a concern inherent to modular pipelines. We provide failure mode analysis in the Supplementary. Notably, our Multi-View Voting mechanism (Sec.~\ref{sec:mv_voting}) is designed to mitigate error propagation.

\subsection{Ablation Study}
\label{sec:ablation}
To ablate the effectiveness of each component in SandboxVLM, we conduct an extensive ablation study on the SAT Real dataset using GPT-5-mini as the backbone VLM. The results are summarized in Tab.~\ref{tab:ablation}. Specifically, we investigate these research questions:
\begin{enumerate}[label=(\roman*)]
  \item \textbf{Information level}: Could the multi-view prior models provide useful extra information for VLMs in spatial reasoning tasks?
  \item \textbf{Visual modality vs.\ text modality}: To represent the spatial information, which modality, visual or textual, is more effective for VLMs?
  \item \textbf{3D vs.\ 2D modality}: Among the visual representations, is 3D-related representation richer in information or more effective than 2D ones?
  \item \textbf{Visualization details of 3D abstraction}: For different 3D abstract visualization designs, how do they affect the final performance of VLMs?
\end{enumerate}
We designed 8 ablation settings to answer these questions, and the settings are detailed below:
\begin{enumerate}[label=(\arabic*)]
  \item Vanilla VLM: The vanilla GPT-5-mini API.
  \item Scene graph text prompt: we applied an expert model for scene graphs~\cite{cong2023reltr} to generate a json file describing the objects and their relations in the single input image, then feed the json description as the context.
  \item Multi-view images only: we first expand the single input image to a multi-view image set using SEVA~\cite{zhou2025svc}, then directly feed the images to the VLM without any 3D lifting or abstraction.
  \item Rendered point cloud: we lift the multi-view images to 3D point clouds using VGGT~\cite{wang2025vggt}, then render the point clouds to multiple views as the context for VLMs.
  \item 3D coordinate text prompt: we generate a text description of the 3D scene, describing the coordinates of centers of object bounding boxes and their sizes, then feed the text description to the VLM.
  \item Rendered Proxy Points: we render the 3D proxy points obtained from our 3D lifting and clustering module to multiple views as the context for VLMs.
  \item Single Image Sandbox: Our method but without the video generation model as multiview priors, leveraging input image only.
  \item Full SandboxVLM: the full system with multi-view prior, proxy lifting, multi-view voting and clustering, and abstract box rendering.
\end{enumerate}
The details and examples of each design are provided in Fig.~\ref{fig:viz} and in the Supplementary. By conducting these ablations, we obtained the following observations:

\begin{figure}[t]
  \centering
  \includegraphics[width=0.95\columnwidth]{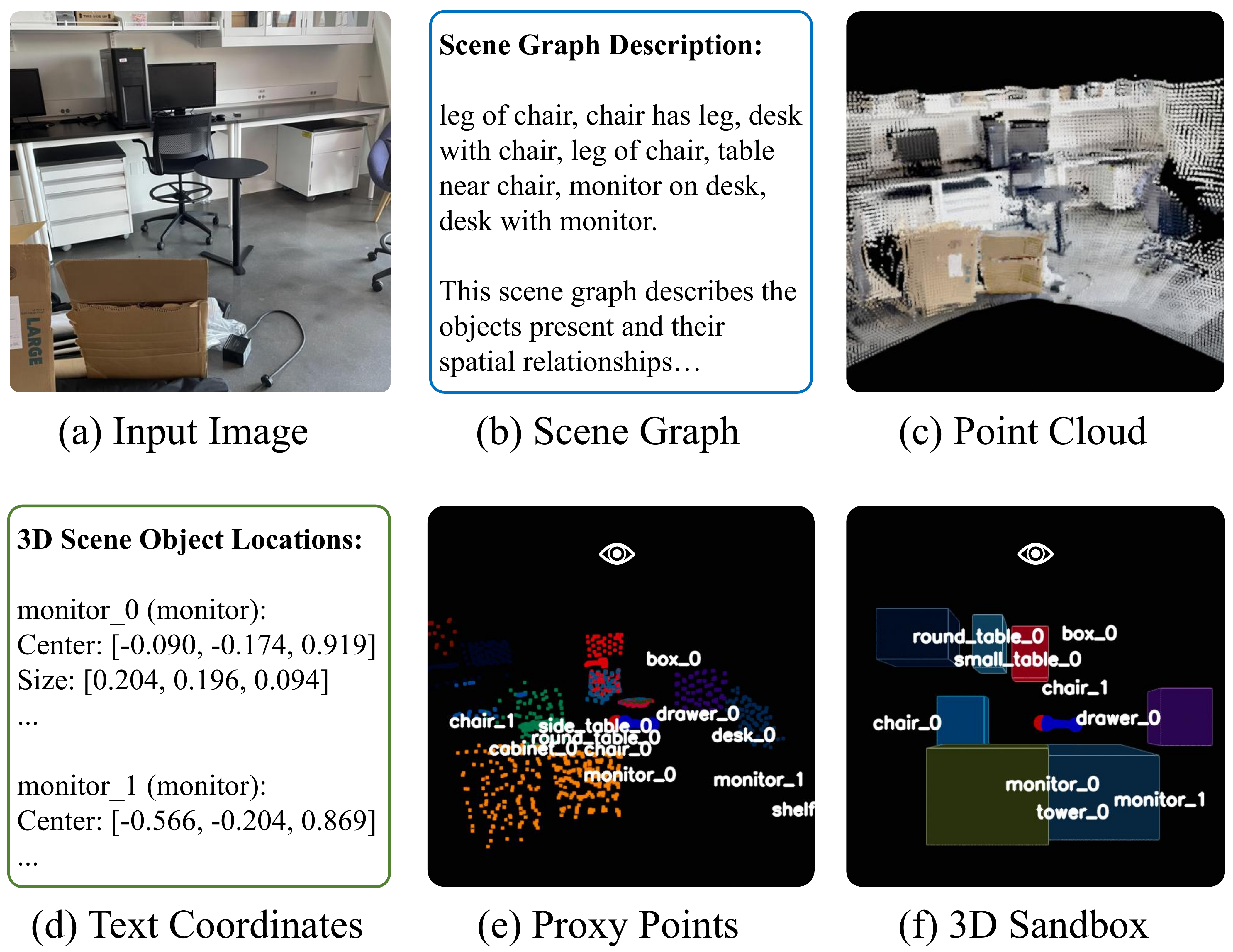}
  \caption{Visualization of representations in ablation study. (a) Input image to the system; (b) Scene graph generated by expert model; (c) Reconstructed point cloud rendering. (d) Text description of 3D bounding boxes; (e) Rendered proxy points; (f) 3D Sandbox. 3D Sandbox strikes a balance between informativeness and interpretability, providing vivid spatial cues while filtering out irrelevant details.}
  \label{fig:viz}
  \vspace{-1.4em}
\end{figure}

\vspace{0.5em}

\noindent\textbf{(i) Multi-view priors are complementary for VLMs.}
Comparing settings (1) and (3), we observe a notable 3.3\% performance boost when incorporating multi-view prior images. (8) outperforms (7) by 6.5\%, indicating that additional viewpoints provide crucial spatial cues that enhance the VLM's reasoning capabilities. This highlights a gap in current generalist VLMs: they lack sufficient 3D spatial knowledge that generative world models inherently possess. The latter can therefore serve as a valuable source of 3D priors to complement VLMs, suggesting a promising direction for future research.

\vspace{0.7em}

\noindent\textbf{(ii) Visual modality is not necessarily more informative to VLMs than text.}
By comparing setting (2) with (3), and (5) with (6), we find that for equivalent information content, 2D images are only marginally more effective (by 1.7\%) than textual descriptions, while 3D box coordinates even outperform rendered proxy points. This suggests that despite their visual capabilities, today's VLMs remain fundamentally language-centric: they struggle to fully exploit rich visual information for complex reasoning. 

\vspace{0.7em}

\noindent\textbf{(iii) Providing 3D information to VLMs benefits spatial intelligence.}
By comparing settings (2) with (5) and (8), we observe that introducing 3D-related information consistently improves spatial reasoning. Setting (5) achieves 80.8\% accuracy, a 3.8\% increase over (2), while (8) further boosts performance to 84.1\%, demonstrating the clear advantage of incorporating external 3D cues over purely 2D representations. However, directly rendering point clouds (setting (4)) results in a substantial performance drop, even below the vanilla baseline, indicating that noisy or sparse 3D inputs can be detrimental. This finding underscores 3D input modality as one of the core questions in developing robust 3D-aware VLMs.

\vspace{0.7em}

\noindent\textbf{(iv) Abstract box rendering is an effective way to convey 3D information to VLMs.}
Comparing settings (4) with (5), (6), and (8), we find that abstracting information from point clouds significantly benefits VLM performance. In particular, setting (6) surpasses (4) by 3.3\%, likely because abstraction filters out irrelevant details present in raw 3D data. Between (5), (6), and (8), directly rendering proxy points (6) proves less effective than providing box coordinates (5) as text input, since rendering can obscure precise spatial details. Meanwhile, setting (8) performs best because abstract bounding boxes retain 3D information through vivid spatial cues and improve the visual clarity of object relations, which explains its advantage over (5). In summary, abstract box rendering offers a strong balance between information richness and interpretability, making it a highly effective means of conveying 3D spatial information to VLMs.

\section{Conclusion}
\label{sec:conclusion}
In this work, we introduced SandboxVLM, a simple yet powerful framework that equips vision-language models with coarse 3D awareness through abstract perception. Rather than relying on dense geometric reconstruction or heavy supervision, our approach injects symbolic 3D structures into existing VLMs via proxy elevation and multi-view voting and clustering—allowing models to reason spatially in a lightweight, training-free manner. Extensive experiments across spatial reasoning benchmarks demonstrate consistent and substantial gains over both generalist and training-based baselines, confirming that abstract 3D representations can meaningfully enhance spatial intelligence.
Beyond bridging the gap between 2D models and 3D reasoning tasks, our findings reveal a broader insight: coarse, interpretable structure is sometimes sufficient for spatial cognition, echoing how humans reason about the physical world. We believe this principle opens new possibilities for scalable, plug-and-play geometric priors that empower future general-purpose VLMs and embodied agents.

\noindent\textbf{Limitations.} (1) Dynamic Scene Adaptability: Our current framework reconstructs static 3D Sandboxes out of each image. While effective for many scenarios, this limits adaptability to dynamic scenes with moving objects or changing layouts, for no correlation between different image views is explicitly modeled in the 3D sandbox.
Future work could explore integrating temporal information or motion cues to better handle dynamic environments. (2) Limited Physical Representation: While our 3D Sandbox captures spatial information effectively, it does not explicitly model physical properties such as mass, friction, or material characteristics. This constrains the depth of physical reasoning possible. Future extensions could incorporate lightweight physical simulations or learned priors from physics expert models to enrich the representation. 

\section*{Acknowledgements}
This work was partially supported by NIH grant R01HD104969, NIH grant 1U01CA284207, and NSF grant CRCNS-2309041.

{
    \small
    \bibliographystyle{ieeenat_fullname}
    \bibliography{main}
}

\clearpage
\setcounter{page}{1}
\maketitlesupplementary

\section{Implementation Details} 
\subsection{Hyperparameters}
We implement SandboxVLM using GPT-5-mini as the backbone VLM, and for variants, we also test with GPT-4o and GPT-5. The results are from the OpenAI API as of September 2025 to November 2025. For abstract control, we allow the VLM to choose from: Left, Forward-left, Forward, Forward-right and Right, which are later converted to trajectories among (1) turning left or right $90^\circ$ (2) moving forward 0.5m then turning left or right $36^\circ$ (3) moving forward 1.0m. For multi-view priors, we use SEVA~\cite{zhou2025svc} to generate 4 additional views per selected trajectory. We utilize the pointmap from VGGT~\cite{wang2025vggt} as the 2D-to-3D lifting model, and the number of views for Multi-View Voting N is set to 3 to balance accuracy and point numbers. In our implementation, $M=3$, $T=8$, $N=2$, $\delta=0.5m$.

\subsection{3D Reasoning Prompt}
\label{suppl:prompt}
We perform 3D reasoning on the rendered images from 3D Sandbox. The prompt is shown as follows: 

\lstset{
    backgroundcolor=\color{codebg},
    basicstyle=\ttfamily\footnotesize,
    breaklines=true,
    breakatwhitespace=true,
    postbreak=\raisebox{0ex}[0ex][0ex]{},
    frame=none,
    columns=fullflexible
}

\begin{lstlisting}[caption={System prompt for spatial reasoning tasks.}, label=code:prompt]
sys_prompt = (
    "You are an AI assistant designed to help with spatial reasoning in a 3D indoor scene. "
    "You must analyze any provided images or observations and answer the question.\n\n"
    "Rules:\n"
    "1. You should output the exact answer from the choices.\n"
    "2. You will be provided with multiple imagined views if you taking corresponding actions to help you answer the questions.\n"
    "3. Your response should be formatted like <think>your analysis for the question and the images, think step by step.</think><answer>your exact answer selected from the choices, do not include other words</answer> Your score will depend on the answer, but thinking clearly and make use of the clues and common sense can help you improve your answer.\n"
\end{lstlisting}

This simple prompt design guides the VLM to systematically perceive and reason about the provided question and 3D context from 3D Sandbox.

\subsection{Subsampling for SAT-Synth}
Since SAT-Synth has 4000 data samples, to control the cost with the OpenAI api, we subsample 98 examples from it, an amount comparable with the 150 samples in SAT-Real split. The subsampled items are listed as follows:

\begin{lstlisting}[caption={The indices of subsampled tasks from SAT-Synth.}, label=code:indices]
val: 1023  1177  13    1407  1727  1932  2057  2279  248   2790  299   3170  33    3437  3790  3952  480  571  657  788 1039  1218  1321  1418  1793  1934  2062  2327  2510  2825  3027  3181  3324  3652  3827  3964  481  619  662  817 1054  1244  1351  1523  1885  2001  2114  2433  2541  2826  3041  3184  3336  3667  3873  3978  509  623  667  93 1076  1261  1357  1559  190   2027  214   2466  259   283   3051  3241  3362  3707  3939  41    514  635  691  results.json 113   1272  1397  1717  1900  2035  2157  2476  2666  2954  3152  3257  3381  3718  3947  416   544  639  761
\end{lstlisting}

\subsection{Details of Benchmarks}
We present the examples of the benchmarks involved in our experiments in Table~\ref{tab:benchmark}, for readers' information.

\subsection{Ablation Design}
\label{suppl:ablation_design}

During all our ablation studies listed in~\ref{sec:ablation}, we observe the prompting method presented in Sec.~\ref{sec:3d_reasoning} and Sec.~\ref{suppl:prompt} to compare the representation side of the design fairly. We further detail the composition of each ablation study experiment in Table~\ref{tab:details_ablation}.

\begin{table*}[!ht]
    \footnotesize\centering
    \caption{Question types and examples from each benchmark. The four types of benchmarks we evaluated on contain various tasks from metric questions like relative depths to comprehensive ones like physical scene understanding. For more details, please refer to the corresponding paper.}
    \begin{tabular}{rp{360pt}}
        \toprule
        \multicolumn{2}{l}{\textit{\textbf{SAT Bench}}~\cite{ray2024sat}} \\
        \midrule
        Ego Movement     & How did the camera likely rotate when shooting the video? \\
        \midrule
        Object Movement & Were any of the objects in the initial frame that you can still see in the second frame moved from their original positions? \\
        \midrule
        Goal Aim & I need to go to the houseplant. Which direction should I rotate to face it? \\
        \midrule
        Action Consequence & I need to go to the mirror (marked 1). If I turn left by 90 degrees, will I be facing away from the mirror? \\
        \midrule
        Perspective & For someone at the x marked point and facing 90 degrees to the left, will the person surfing be to their left or right? \\
        \midrule
       \multicolumn{2}{l}{\textit{\textbf{BLINK}}~\cite{fu2024blink}} \\
        \midrule
        Relative Depth & Which point is closer to the camera? \\
        \midrule
        Spatial Relation & Is the bed at the right side of the dining table? \\
        \midrule
        \multicolumn{2}{l}{\textit{\textbf{EmbSpatial Bench}}~\cite{du2024embspatialbench}} \\
        \midrule
        Spatial Relation     &How are statue and tissuebox positioned in relation to each other in the image? (multiple choices) \\
        \midrule
        & A. The statue is at the left side of the tissuebox. \\
        \midrule
        & B. The statue is touching the tissuebox. \\
        \midrule
         & C. The statue is outside the tissuebox.\\
        \midrule
         & D. The statue is right of the tissuebox. \\
                 \toprule
        \multicolumn{2}{l}{\textit{\textbf{PhysBench}}~\cite{chow2025physbench}} \\
        \midrule
        Physical Object Property     & Which color of balls 
has the largest number? \\
        \midrule
        & Given that the applied 
force is the same, which 
object in the images has 
higher stiffness? \\
        \midrule
        Physical Object Relationships & What is the distance between 
the yellow cube and the blue ball? 
(The blue cube has a width of 2 
cm.) \\
        \midrule
         & Which car has a higher average speed?  \\
        \midrule
        Physical Scene Understanding & How does the focal 
length of the camera 
change? \\
\midrule
        & What causes the change in 
water level in the cup? \\
        \midrule
        Physical-based Dynamics & Which can is the ball 
most likely to land in? \\
\midrule
        & What is the correct sequence 
of images to make a gift box 
containing the perfume bottle? \\
        \bottomrule
    \end{tabular}

    \label{tab:benchmark}
\end{table*}

\begin{table*}[t]
\centering
\vspace{3em}
\caption{Composition of each ablation method. We list: Average: the average performance on SAT-Real~\cite{ray2024sat}; Modality: in which modality is the context information provided to VLM; Priors: whether a world model is used as a 3D-aware prior in generating the context; 3D: whether the context is lifted into the 3D space; Filtered: before constructing the context, whether a VLM is used to collect information relevant with the task; Using Boxes: whether we finally construct the 3D Sandbox as the context.}
\label{tab:details_ablation}
\setlength{\tabcolsep}{6pt}
\begin{tabular}{c l *{6}{c}}
\toprule
No. & Method & Average & Modality & Priors & 3D & Filtered &Using Boxes\\
\midrule
(1) & Vanilla VLM & 75.4& -& No& No&No& No\\
(2) & Scene-Graph Text Prompt & 77.0& Text& Yes& No&No& No\\ 
(3) & Multi-View Images Only & 78.7& Vision & Yes& No&No& No\\
(4) & Rendered Point Clouds & 73.7& Vision& Yes& Yes& No& No\\
(5) & 3D Coordinate Text Prompt& \underline{80.8}& Text& Yes& Yes&Yes& Yes \\
(6) & Rendered Proxy Points & 77.0& Vision& Yes& Yes& Yes&No\\
\rowcolor{gray!15}
(7) & Full SandboxVLM    & \textbf{84.1} & Vision  & Yes & Yes  &Yes& Yes\\
\bottomrule
\end{tabular}
\end{table*}

\section{Error Analysis}
Error propagation is inherent to modular pipelines. In our framework, error sources include video diffusion, VLM pointing, image segmentation, and depth/camera estimation. We provide qualitative visualizations and human-verified statistics to analyze these errors.
\begin{figure*}[h]
  \centering
  \includegraphics[width=0.7\textwidth]{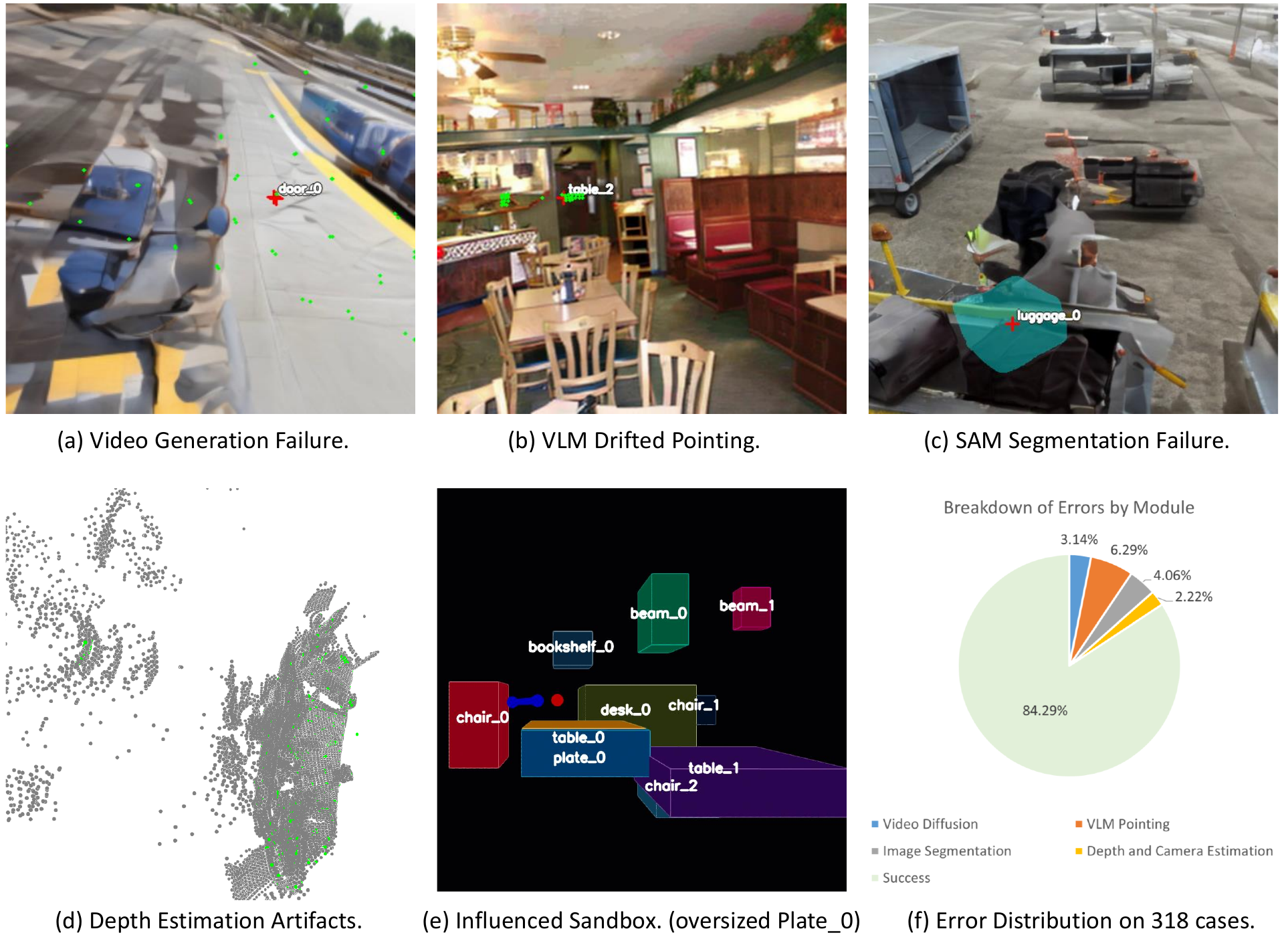}
  \caption{(a)-(e): Different modes of failure cases. (f) Error distribution statistics on SAT-Real. Please zoom in for a better view.}
  \label{fig:ablation}
\end{figure*}

Most failures stem from inaccurate VLM pointing, which often propagates to or reinforces image segmentation errors. Imprecision in video generation and depth or camera estimation primarily introduces dispersed 3D point noise. Our abstraction and the Multi-View Voting mechanism (Section 3.4) are explicitly designed to suppress the propagation of these errors and imprecisions.

\section{More Analysis}

\subsection{SAT Benchmark}
Across different backbones on SAT-Real, we observe consistent gains in the Ego-Movement, Goal-Aiming, Action-Consequence, and Perspective subtasks, with particularly strong results on Ego-Movement (exceeding 90\% accuracy and reaching 100.0\% in some settings). We attribute these improvements to the explicit 3D spatial representation and reasoning process in our framework, which effectively captures view changes within a stable and coherent 3D context. For Perspective tasks—where the model must reason from another agent's or object's viewpoint—SandboxVLM also achieves notable gains, benefiting from its ability to reconstruct only relevant regions of the 3D scene and represent objects via 3D-aware bounding boxes.

The relatively lower performance on the Object-Movement subtask compared with MindJourney, GPT-4o as backbone on SAT-Synth, likely stems from the static nature of our current 3D reconstruction module. Extending SandboxVLM toward dynamic 3D reconstruction would thus be a promising direction for future research. 

\begin{figure*}[t]
  \centering
  \includegraphics[width=0.83\textwidth]{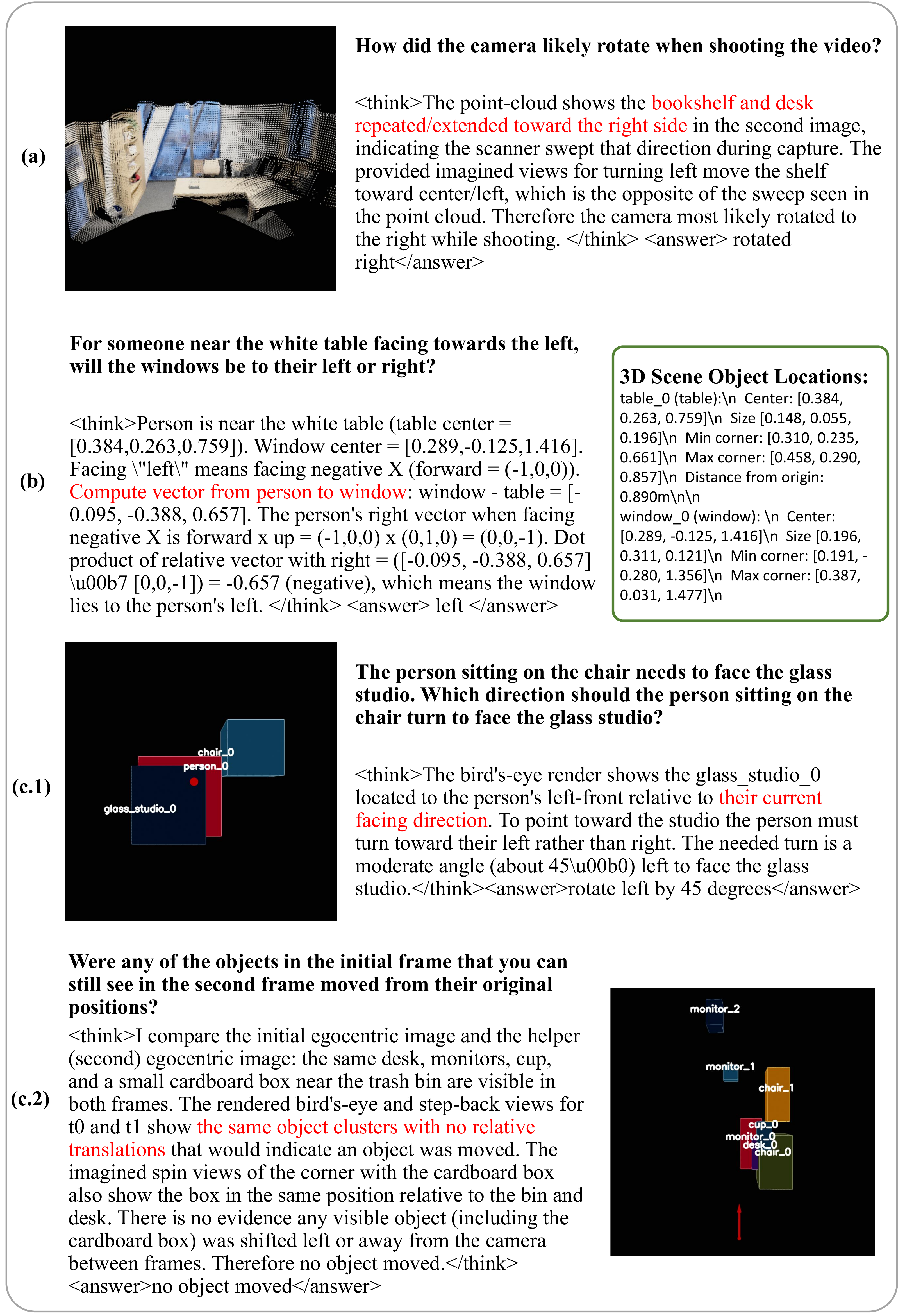}
  \caption{\textbf{Examples of failure cases.} We show failure cases from different types of inputs: (a) Rendered Point Clouds; (b) 3D Coordinate Text Prompt; (c) Ours. The erroneous messages are highlighted in red. We provide analysis of these failure modes in Sec.~\ref{suppl:failure_case}.}
  \label{fig:failure_case}
\end{figure*}

\subsection{Representation Analysis}
\label{suppl:failure_case}
We provide analysis of failure cases from different types of inputs in Fig.~\ref{fig:failure_case}.

\noindent\textbf{(a) Rendered Point Clouds.} The sparse and incomplete nature of the point cloud representation can lead to misinterpretation of spatial relationships. In the example, the VLM misjudges that the bookshelf also apperas on the right side, causing confusions in interpreting the spatial layout. This flaw highlights the limitations of relying solely on point cloud rendering for spatial reasoning tasks, pointing to a gap between the 3D reconstruction research and VLMs, a gap which our work aims to bridge.

\noindent\textbf{(b) Scene-graph Text Prompt.} The 3D coordinate text prompt provides filtered and structured information about object locations in the text format. However, in most cases we observed that the VLM began reasoning in pure text format with the provided numerical coordinates. It even started computing vectors and cross products to deduce spatial relationships, without attending to visual signals. We attribute this to the strong bias towards the language modality in VLM CoT, where the model tends to prioritize textual reasoning over visual cues when both are present. This also validates our methods' motivation in constructing a visual 3D context for the VLM to perceive.

\noindent\textbf{(c) Ours.} We also analyze two typical failure modes in our framework: lack of appearance details, and lack of modeling dynamics. In (c.1), we demonstrate a case where the question involves the original facing direction of the human inside the scene. However, our 3D Sandbox does not model the exact initial facing direction of the human, leading to hallucinations. This is due to our focus on geometric cues rather than appearance in the representation. Future work addressing appearance modeling in 3D Sandbox, e.g. with Sam-3D-Objects, would help mitigate this issue. In (c.2), the VLM fails to correctly discern the object movement between two 3D Sandbox renderings. We attribute this to (i) the errors introduced in the reconstruction pipeline such as VLM pointing and the 3D reconstruction from generated views, which can misalign objects slightly; (ii) the static nature of our current 3D reconstruction module, which cannot model dynamic changes directly in one 3D Sandbox. Addressing these limitations would be a promising direction for future research.

\end{document}